\documentclass[pmlr,twocolumn,10pt]{jmlr} 

\mlhtrack{findings}

\newif\iffinal
 \finaltrue  

\iffinal
    \ifmlhneedspmlr
      \jmlrvolume{XXX}
      \jmlryear{2026}
    \fi
    \ifmlhfindings
      \jmlrproceedings{}{}
    \fi
\fi

\usepackage{booktabs}
\usepackage{siunitx}

\title[Lifestyle Information and Hypertension Prediction]{Does Lifestyle Information Improve Hypertension Prediction? A Leakage-Controlled Evaluation on NHANES}

\author{\Name{Uriel Benymon} \Email{udb2101@columbia.edu}\\
\addr Columbia University, New York, NY, USA}

\begin{document}

\maketitle

\ifmlhdemo\else
\begin{abstract}
Does readily available lifestyle information improve machine-learning
prediction of hypertension beyond routine demographic and
anthropometric data? On NHANES 2017--March 2020, five model families
(Logistic Regression, Random Forest, Gradient Boosting, Bagged Decision
Trees, and XGBoost) were evaluated with fold-local preprocessing and
paired bootstrap confidence intervals. Adding physical activity,
alcohol use, smoking, and sleep information showed no consistent
detectable improvement in held-out discrimination: every paired ROC-AUC
confidence interval included zero, including under a stricter
measured-blood-pressure phenotype and without synthetic oversampling.
Mean cross-model SHAP rank agreement fell from 0.873 to 0.806 over all
encoded columns, but this decline was largely attenuated over the 19
baseline columns shared by both arms (0.873 vs.\ 0.863), and likewise
under both robustness checks; attribution-agreement comparisons across
feature sets of different sizes therefore require a matched comparison
space. In exploratory analysis of untreated adults
(measured-blood-pressure outcome), all five models showed small
positive differences, two with unadjusted intervals excluding zero,
meriting confirmation. These results do not imply that lifestyle
factors are clinically unimportant; within this cohort, the evaluated
lifestyle block added little incremental discrimination beyond an
already-informative baseline.
\end{abstract}

\fi

\ifmlhneedsstatements
\paragraph*{Data and Code Availability}
This study uses publicly available, de-identified NHANES 2017--March 2020
data from NCHS/CDC, freely accessible at
\url{https://wwwn.cdc.gov/nchs/nhanes/continuousnhanes/default.aspx?Cycle=2017-2020}.
All code, result tables, and figures are publicly available at \url{https://github.com/urielbenymon/nhanes-lifestyle-hypertension}, including scripts that download the NHANES data directly from CDC and reproduce every reported analysis.

\paragraph*{Institutional Review Board (IRB)}
This study is a secondary analysis of publicly available, de-identified
NHANES data; no new data were collected and no interaction with human
subjects occurred. No IRB review was sought for this analysis.
\fi

\section{Introduction}
\label{sec:intro}

Hypertension affects a large share of the global adult population and
remains a leading modifiable contributor to cardiovascular disease,
with sustained increases in prevalence and shifting burden documented
across more than a hundred million participants worldwide
\citep{zhou_worldwide_2021}. The 2017 ACC/AHA guideline lowered the
diagnostic threshold to 130/80~mmHg, expanding the population
classified as hypertensive and increasing interest in identifying who
is at risk from readily available information
\citep{whelton_2017_2018}. Physical activity, alcohol use, smoking, and
sleep are frequently recommended targets for hypertension prevention
and management, making it practically useful to know how much they add
to prediction built from routinely collected demographic and
anthropometric data.

A substantial body of work applies machine learning to hypertension
prediction, spanning multiple model families, feature sets, and
evaluation designs \citep{silva_machine_2022}. Closest to the present setting, \citet{potvin_using_2025}
predict hypertension from NHANES demographic
and lifestyle variables using Logistic Regression, Random Forest,
Gradient Boosting, and Bagged Decision Trees with a single 90/10
evaluation split; \citet{huang_shapely_2023} predict hypertension from
the same NHANES 2017--March 2020 cycle using XGBoost and SHAP over
demographic and nutritional covariates; and
\citet{lopez-martinez_artificial_2020} predict hypertension from an
earlier NHANES cycle using a neural network. Collectively, this work
establishes that demographic, anthropometric, clinical, nutritional,
and behavioral information can each participate in hypertension
prediction, though evaluation designs, phenotype definitions, and
cohorts vary considerably across studies, which complicates direct
comparison of reported performance.

Physical activity, alcohol use, smoking, and sleep are each
independently associated with hypertension risk
\citep{barone_gibbs_physical_2021, hosseini_association_2024,
cecchini_alcohol_2024, virdis_cigarette_2010}. However, established
risk-factor association is a distinct question from incremental
predictive value: a variable can be epidemiologically relevant to
hypertension while adding little additional discriminative information
once other readily available predictors are already present. It
remains less clear how much a focused, modifiable lifestyle block
contributes beyond an already-informative demographic and
anthropometric baseline under leakage-controlled evaluation, and
whether expanding the feature space in this way changes how
consistently different model families identify important predictors, a
question distinct from whether any single model's attribution is
trustworthy in isolation.

This gap is addressed using the NHANES 2017--March 2020 cycle,
comparing a demographic/SES-and-anthropometric baseline against a
lifestyle-augmented arm across five model families
(Section~\ref{sec:methods}). The first empirical question (EQ1) is whether the lifestyle block
provides incremental held-out predictive value beyond the baseline. The
second empirical question (EQ2) is whether it changes cross-model agreement in SHAP
feature importance, and whether any change persists when both arms are
compared over the same baseline columns rather than over feature spaces
of different sizes. Robustness is tested under an alternative
hypertension phenotype and removal of synthetic oversampling
(Appendix~\ref{apd:robustness}), supplemented by calibration and
decision-curve analysis, feature-subset ablations, training-seed and
repeated-split variability, an untreated-adult subgroup, and
lifestyle--baseline redundancy diagnostics.

Relative to \citet{potvin_using_2025}, the contribution lies mainly in
the evaluation design: an explicit incremental-value comparison with
paired feature arms on the same participants and split; fold-local
preprocessing, including SMOTE, with cross-validated tuning; paired
held-out uncertainty for the lifestyle-minus-baseline difference;
XGBoost as a fifth model family; and a cross-model attribution analysis
with robustness checks. Two further findings emerge. First, cross-model
SHAP agreement compared across feature sets of different sizes is
sensitive to how the comparison space is defined, so such comparisons
require a matched comparison space. Second, the full-cohort null result
may not hold in every subpopulation: in an exploratory analysis of
untreated adults, all five models showed small positive
lifestyle-minus-baseline AUC differences, a result presented as
hypothesis-generating.
\section{Methods}
\label{sec:methods}

\textbf{Cohort and outcome.} This study uses the NHANES 2017--March 2020
pre-pandemic release, restricted to adults aged 18 and older who were
both interviewed and examined. Of 15{,}560 initial participants, 8{,}965
met this criterion, and 8{,}024 additionally had a valid oscillometric
blood pressure reading and were phenotype-eligible (Table~\ref{tab:cohort}).
Hypertension was defined using the 130/80~mmHg threshold established by
\citet{whelton_2017_2018}, averaged across up to three readings per
participant. The primary outcome, Phenotype B, is positive if measured
blood pressure meets this threshold or the participant reports
currently taking antihypertensive medication (4{,}311/8{,}024, 53.7\%);
the medication clause is not redundant with measured BP alone, since
903 of these positives have normal measured blood pressure. A
sensitivity analysis, Phenotype A, uses the measured-BP criterion alone
(3{,}408/8{,}024, 42.5\%). Full phenotype definitions, including a
broader self-report-inclusive Phenotype C, and the complete predictor
dictionary appear in Appendix~\ref{apd:cohort}. Participants were split
once into a fixed 90\% development set (7{,}221) and a 10\% held-out
test set (803), stratified on Phenotype B; this split was reused
unchanged for the Phenotype A analysis so that only the outcome label,
not cohort composition, varied. Survey design variables were retained
but not used as predictors, so reported estimates describe predictive
performance within this analytic cohort rather than
population-representative prevalence.

\textbf{Features and modeling.} Two feature arms are compared using the
same participants and split: a \emph{baseline} arm (demographic/SES and
anthropometric predictors) and a \emph{lifestyle-augmented} arm that
adds physical activity, alcohol use, smoking, and sleep. This block was
prespecified as commonly recommended modifiable behaviors from the
guideline and meta-analytic evidence cited in Section~\ref{sec:intro},
rather than selected using the outcome, which would tune it to the data
it is tested on and bias the incremental comparison; every candidate
variable has a documented inclusion or exclusion decision, made from
the study aims and cited literature without formal clinical expert
review (Appendix~\ref{apd:cohort}). Twenty-six raw lifestyle items were
consolidated into 11 derived predictors using skip-pattern-aware rules.
This pairing is the instrument for the first empirical question: whether the
lifestyle block adds incremental predictive value beyond an
already-informative baseline. Each arm was evaluated with five model
families: Logistic Regression, Random Forest, Gradient Boosting, Bagged
Decision Trees, and XGBoost \citep{chen_xgboost_2016}. Every pipeline
applies one-hot encoding, median imputation, standardization for
Logistic Regression only, and SMOTE \citep{chawla_smote_2002}, all fit
inside an \texttt{imbalanced-learn} pipeline
\citep{lemaitre_imbalanced-learn_2016} so that no learned
transformation, including SMOTE's oversampling, was ever fit on
validation or test data, addressing a documented source of
leakage-driven irreproducibility in machine-learning-based research
\citep{kapoor_leakage_2023}. Hyperparameters were selected via 5-fold
stratified cross-validation \citep{kohavi_study_1995} on the 90\%
development set only, scored by ROC-AUC: an exhaustive grid for
Logistic Regression, and 25-candidate randomized search
\citep{bergstra_random_2012} over tree depth, ensemble size, and
related regularization parameters for the remaining four models, run
independently per model per feature arm, ten configurations total,
with full search spaces and selected values in
Appendix~\ref{apd:training}. This is a single cross-validation layer,
not a nested search. Selected classifier hyperparameters were frozen
and reused, unmodified, for held-out evaluation, interpretability
analysis, both robustness checks, and all supplementary analyses below.

\textbf{Evaluation.} ROC-AUC and PR-AUC are reported. Held-out
performance uses bootstrap 95\% confidence intervals
\citep{efron_introduction_1993} on the fixed 10\% test partition. In
the main-model evaluation, the test set was examined twice: once with
library-default hyperparameters and once after all hyperparameters were
frozen; it was never used to fit models or select hyperparameter values. The default
results contributed to the decision to tune, but that decision was
supported by development data alone: XGBoost's default
cross-validated ROC-AUC (0.782 baseline, 0.785 lifestyle-augmented) was
already well below that of Logistic Regression (0.813, 0.815). All five
models were then tuned with the same development-set criterion, not
XGBoost alone, and no search space, feature set, or preprocessing
decision was changed in response to held-out results. The default run
had already produced the same null result for the first question (all
five paired ROC-AUC intervals included zero; point estimates $-$0.014
to $+$0.010). The final test evaluation is nonetheless not fully
independent of earlier modeling decisions, and the supplementary
analyses below reuse the same test set.

To test the first question, a paired bootstrap confidence interval is
computed for the lifestyle-minus-baseline difference, resampling the
same test participants for both arms each draw
(Figure~\ref{fig:test_delta}; full values in
Appendix~\ref{apd:results}). For the second question, whether adding
lifestyle information changes how consistently model families
attribute importance, SHAP values are computed
\citep{lundberg_unified_2017} using \texttt{TreeExplainer}
\citep{lundberg_local_2020} for Random Forest, Gradient Boosting, and
XGBoost, and a model-agnostic explainer for Bagged Decision Trees,
which \texttt{TreeExplainer} does not support; these describe learned
attribution, not a causal claim about hypertension risk. Following
\citet{krishna_disagreement_2022}'s framing of cross-model explanation
disagreement, agreement is quantified two ways: top-10 Jaccard overlap
and Spearman rank correlation over the complete ordering, both over the
post-encoding feature space and, because the two arms rank different
numbers of columns (19 vs.\ 36), also over the 19 baseline columns they
share (Table~\ref{tab:shap_stability};
Appendix~\ref{apd:interpretability}). Robustness is tested by refitting
under Phenotype A and by refitting with SMOTE removed, holding
classifier hyperparameters fixed (Appendix~\ref{apd:robustness}).

Supplementary analyses, all using the frozen hyperparameters, comprise
SHAP agreement controls (matched comparison space, a shuffled-lifestyle
control, and predictor- and domain-level aggregation;
Appendix~\ref{apd:shapcontrols}); calibration and exploratory
decision-curve analysis (Appendix~\ref{apd:calibration});
lifestyle--baseline redundancy diagnostics on the development set
(Appendix~\ref{apd:redundancy}); an untreated-adult subgroup
(Appendix~\ref{apd:untreated}); feature-subset ablations
(Appendix~\ref{apd:ablation}); and training-seed and repeated-split
variability (Appendix~\ref{apd:variability}). Reproducibility and
ethics details appear in Appendices~\ref{apd:repro}
and~\ref{apd:ethics}.

\begin{table*}[htbp]
\floatconts
  {tab:cohort}
  {\caption{Cohort construction, phenotype definitions, and feature-arm
  composition. Phenotype C is not analyzed in the main text; see Appendix~A.}}
  {\begin{tabular}{lr}
  \toprule
  \bfseries Item & \bfseries Value \\
  \midrule
  Initial NHANES 2017--March 2020 participants & 15{,}560 \\
  Interviewed and MEC-examined & 14{,}300 \\
  Age $\geq$ 18 & 8{,}965 \\
  Phenotype-eligible (valid BP reading) & 8{,}024 \\
  \midrule
  Primary Phenotype B (measured $\geq$130/80 or medication) & 4{,}311 (53.7\%) \\
  Sensitivity Phenotype A (measured only) & 3{,}408 (42.5\%) \\
  \midrule
  Development / held-out test split & 7{,}221 / 803 \\
  \midrule
  Baseline predictors & Demographic/SES + anthropometric \\
  Lifestyle-augmented predictors & Baseline + activity, alcohol, smoking, sleep \\
  \bottomrule
  \end{tabular}}
\end{table*}

\section{Results}
\label{sec:results}

Table~\ref{tab:cohort} summarizes the analytic cohort and both feature
arms. After cross-validated hyperparameter selection, lifestyle
augmentation produced only small, inconsistent changes in
development-set discrimination across the five model families, with
ROC-AUC differences ranging from $-$0.0005 for Random Forest to
$+$0.0035 for Gradient Boosting; Random Forest is the only model with a
negative development-set delta. These are development/model-selection
estimates rather than an independent confirmation, since the same
cross-validation folds informed hyperparameter selection.
 
On the fixed held-out test set, lifestyle augmentation did not reliably
improve discrimination for any model family
(Figure~\ref{fig:test_delta}). Paired bootstrap ROC-AUC differences
ranged from $-$0.0028 for XGBoost to $+$0.0002 for Bagged Decision
Trees, and all five 95\% confidence intervals included zero; the
corresponding PR-AUC differences showed the same pattern, with full
values in Appendix~\ref{apd:results}. Small development-set differences did not translate into reliable
held-out improvement. This result addresses
incremental predictive discrimination specifically, within this
analytic cohort; it does not speak to whether lifestyle factors are
associated with hypertension more broadly.

\begin{figure}[htbp]
\floatconts
  {fig:test_delta}
  {\caption{Paired bootstrap comparison of held-out ROC-AUC,
  lifestyle-augmented minus baseline, tuned models (95\% CI). All five
  intervals include zero.}}
  {\includegraphics[width=\linewidth]{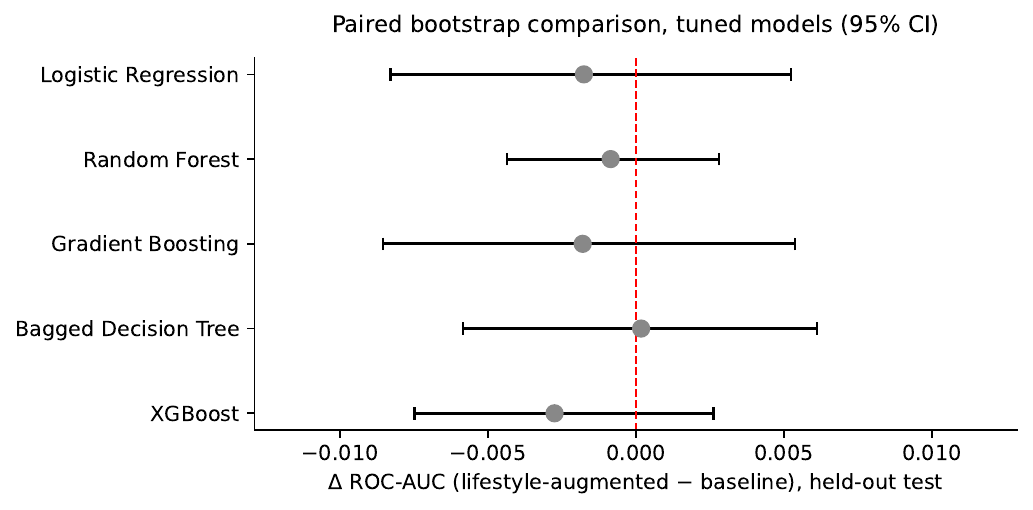}}
\end{figure}

At the encoded-column level, mean pairwise cross-model SHAP rank
agreement decreased after lifestyle augmentation (Spearman 0.873 to
0.806; Table~\ref{tab:shap_stability}). However, the decrease was
substantially attenuated when both arms were compared on the same 19
baseline columns (0.873 vs.\ 0.863; paired bootstrap $\Delta = -0.010$,
95\% CI [$-0.021$, 0.010]). A shuffled-lifestyle control, which
preserves the 36-column feature space and each lifestyle predictor's
marginal distribution while disrupting its observed associations, showed
a similar full-column decline (0.776) with similar shared-column
agreement (0.863). Top-10 Jaccard showed a smaller lifestyle-specific
component: it fell further with real lifestyle columns (0.699) than with
shuffled ones (0.745), as real lifestyle features such as trouble
sleeping entered some models' top 10 but not others. Shuffled columns
received 5.3\% of total mean $|$SHAP$|$, compared with 11.3\% for real
lifestyle columns. Predictor- and domain-level analyses indicated
substantial agreement at coarser resolutions, subject to their
aggregation and unequal-list-length limitations: summed within six
domains, mean pairwise Spearman agreement in the lifestyle arm was
0.903, and every model ranked demographic/SES first and anthropometric
second (Appendix~\ref{apd:shapcontrols}). The full-column comparison is
therefore sensitive to how the feature space is defined, and it should
not be interpreted as a general deterioration in cross-model
explanation agreement. All measures describe each model's learned
attribution, not a causal ranking of hypertension risk factors, and are
interpreted descriptively; an illustrative single-model predictor-level
ranking appears in Appendix~\ref{apd:interpretability}.

\begin{table*}[htbp]
\floatconts
  {tab:shap_stability}
  {\caption{Mean pairwise cross-model SHAP agreement over encoded
  columns, primary phenotype. ``All'' compares each condition over its
  full column set; ``shared 19'' restricts every condition to the 19
  baseline columns. The shuffled row independently permutes each
  lifestyle predictor within the development and test sets (a single
  realization) and refits all models. Raw top-10 Jaccard values are not
  comparable across list lengths (chance level 0.365 for 19 columns,
  0.167 for 36). $\Delta$ rows give the paired bootstrap difference
  (real lifestyle minus baseline) over the 300 explained participants;
  these intervals do not include model-refit variability.}}
  {\small\begin{tabular}{lccccc}
  \toprule
  & & \multicolumn{2}{c}{\bfseries Spearman $\rho$} & \multicolumn{2}{c}{\bfseries Top-10 Jaccard} \\
  \cmidrule(lr){3-4}\cmidrule(lr){5-6}
  \bfseries Condition & \bfseries Columns & \bfseries All & \bfseries Shared 19 & \bfseries All & \bfseries Shared 19 \\
  \midrule
  Baseline & 19 & 0.873 & 0.873 & 0.821 & 0.821 \\
  Lifestyle-augmented (real) & 36 & 0.806 & 0.863 & 0.699 & 0.761 \\
  Lifestyle-augmented (shuffled) & 36 & 0.776 & 0.863 & 0.745 & 0.788 \\
  \midrule
  $\Delta$ real $-$ baseline & & $-$0.067 & $-$0.010 & $-$0.122 & $-$0.061 \\
  \quad 95\% CI & & [$-$0.081, $-$0.049] & [$-$0.021, 0.010] & [$-$0.189, $-$0.045] & [$-$0.104, 0.030] \\
  \bottomrule
  \end{tabular}}
\end{table*}

The absence of a reliable held-out lifestyle benefit persisted under
two robustness checks (Appendix~\ref{apd:robustness}). Refitting the
frozen configurations under the measured-BP-only Phenotype A left all
five paired confidence intervals including zero, despite Phenotype A's
substantially lower overall discrimination (development-set ROC-AUC
near 0.71--0.72, versus approximately 0.81 for the primary phenotype).
Removing SMOTE while holding classifier hyperparameters fixed produced
held-out ROC-AUC changes of only $-$0.0031 to $+$0.0029, with all ten
paired confidence intervals again including zero. The reinterpreted
SHAP pattern also held under both checks: full-column Spearman
agreement declined (Phenotype A $\Delta = -0.036$ [$-0.053$, $-0.013$];
without SMOTE $-0.069$ [$-0.088$, $-0.059$]), whereas agreement over
the shared 19 baseline columns did not decline (Phenotype A $+0.029$
[$+0.008$, $+0.053$]; without SMOTE $+0.008$ [$-0.017$, $+0.017$];
Appendix~\ref{apd:shapcontrols}).

Supplementary analyses were broadly consistent with the primary result.
Brier scores and calibration were similar across arms, and 19 of 20
exploratory decision-curve net-benefit comparisons had intervals
including zero; the exception, Bagged Decision Trees at a threshold of
0.5, favored the lifestyle arm (Appendix~\ref{apd:calibration}). In
feature-subset ablations, age alone reached held-out ROC-AUC
0.775--0.780 and the full baseline 0.804--0.810, whereas lifestyle
alone reached 0.629--0.648 (Appendix~\ref{apd:ablation}). Across ten
training seeds on the original split, none of 50 paired
lifestyle-minus-baseline intervals excluded zero, although Bagged
Decision Trees showed a small positive difference in all ten seeds
($+$0.0001 to $+$0.0042); exploratory repeated random splits gave
similar results (Appendix~\ref{apd:variability}).

In an exploratory analysis restricted to participants definitely not
taking antihypertensive medication ($n = 5{,}697$; test set 566), with
measured-BP Phenotype A as the outcome, all five models showed positive
lifestyle-minus-baseline ROC-AUC differences ($+$0.0003 to $+$0.018).
Unadjusted intervals excluded zero for Bagged Decision Trees ($+$0.018
[$+$0.002, $+$0.035]) and, narrowly, XGBoost ($+$0.012 [$+$0.00001,
$+$0.024]; Appendix~\ref{apd:untreated}). This analysis changes both
the population and the outcome definition, reuses hyperparameters tuned
for the primary phenotype in the full cohort, has a small test set, and
involves five comparisons not adjusted for multiplicity; it is a
sensitivity analysis, not an estimate of a treatment effect.

\section{Discussion and Conclusion}
\label{sec:discussion}

In the full NHANES analytic cohort, the evaluated lifestyle block of
physical activity, alcohol, smoking, and sleep information showed
little detectable incremental discrimination beyond demographic,
socioeconomic, and anthropometric predictors, across five model
families and under both robustness checks. Exploratory analysis among
definitely untreated adults yielded consistently positive, generally
small AUC differences, with two nominal unadjusted intervals excluding
zero; these findings merit confirmation in independently validated,
appropriately tuned analyses. Because that analysis changes both the
population and the outcome definition, it cannot distinguish among
differences in treatment status, outcome definition, or population
characteristics. None of this is evidence that lifestyle behaviors are
unassociated with hypertension or clinically unimportant; established
risk-factor associations and incremental predictive discrimination in a
specific analytic cohort are different questions, and these results
speak only to the latter.

Descriptive diagnostics on the development set help explain the small
increment (Appendix~\ref{apd:redundancy}). The lifestyle variables are
not duplicates of the baseline: their largest pairwise Spearman
correlation with a baseline predictor was moderate ($|\rho| = 0.30$,
vigorous activity vs.\ age), and the full baseline predicted each
lifestyle variable only weakly. However, many lifestyle--outcome
associations weakened substantially after accounting for
baseline-predicted risk (e.g., vigorous activity, $\rho = -0.16$ to
$-0.05$; former smoking, $0.13$ to $0.01$), with trouble sleeping a
partial exception ($0.14$ to $0.08$). This pattern is consistent with
overlapping predictive information; it does not establish conditional
independence or a causal explanation.

The full-column decline in cross-model SHAP agreement should not be
read as a general deterioration of explanation agreement. Over the 19
baseline columns shared by both arms, the estimated change was small,
with a confidence interval spanning zero, and a single shuffled-lifestyle
control showed a similar full-column decline. This supports, but does
not prove, the interpretation that the full-column comparison is
sensitive to adding low-importance columns that models order
inconsistently. Real lifestyle columns nevertheless received more
attribution than shuffled ones and entered some models' top 10 but not
others, so the remaining disagreement is concentrated among the small
lifestyle features rather than the baseline. The broader methodological
point is that comparisons of attribution agreement across feature sets
of different sizes require a matched comparison space; agreement over
the full encoded space conflates changes in how models rank shared
predictors with the addition of new predictors to rank. Aggregating attributions to predictor or domain level is a useful
complement, since all five models agreed on the leading domains
(Appendix~\ref{apd:shapcontrols}), although such aggregation can mask
opposing within-group contributions. \citet{huang_shapely_2023} similarly found that
demographic factors dominated SHAP-based importance over a nutritional
covariate set in the same NHANES cycle; this finding is a related but
methodologically distinct observation, since theirs concerns
within-model attribution rather than incremental ablation.

Several limitations bound these conclusions. The analysis uses a single
NHANES cycle without external validation; testing whether the pattern
holds in other NHANES cycles or in external cohorts is the main
direction for future work. Estimates describe predictive performance
within this analytic cohort rather than a survey-weighted, nationally
representative population. Lifestyle here refers specifically to the
activity, alcohol, smoking, and sleep block evaluated; diet and
laboratory biomarkers were excluded by design, and a broader block may
behave differently. Variables were selected from the study aims and the
cited literature without a formal clinical expert review. Performance
was not evaluated within sex, age, or race/ethnicity subgroups. The
held-out test set was examined once with default hyperparameters before
tuning, and all supplementary analyses reuse it, so the final
evaluation is not fully independent of earlier modeling decisions
(Section~\ref{sec:methods}). The SMOTE sensitivity result reflects
hyperparameters selected under the SMOTE-inclusive primary pipeline
rather than independently retuned alternatives. Future work could also
examine blood-pressure control among the 2{,}313 participants currently
taking antihypertensive medication, which is a distinct outcome and
population, and other modifiable domains such as diet. For this
application, the baseline predictor set captured nearly as much
discriminative information as the lifestyle-augmented set.

\acks{This work was supported by the Bancroft Research Scholars program at
Columbia University. A special thanks to Professor Ansaf Salleb-Aouissi (Columbia
University) for guidance during the early stages of this project, including the
recommendation to use cross-validation and SHAP-based interpretability.}

\bibliography{ref}

\appendix

\section{Cohort, Phenotype, and Variable Specification}
\label{apd:cohort}

Hypertension was defined using the 130/80~mmHg threshold established by
\citet{whelton_2017_2018}. Three definitions were computed from the same
cohort: (A) measured systolic $\geq$130~mmHg or diastolic $\geq$80~mmHg;
(B, primary) A or currently taking antihypertensive medication; (C) B or
a self-reported hypertension diagnosis. Table~\ref{tab:phenotype_full}
reports all three. Within Phenotype B's 4{,}311 positives, 3{,}408 met
the measured-BP criterion and 903 were medication-only positives (normal
measured BP, on medication) -- confirming Phenotype B's medication clause
meaningfully changes case status. Measured BP was averaged across all
available oscillometric readings (up to three) per participant.

\begin{table}[htbp]
\floatconts
  {tab:phenotype_full}
  {\caption{Prevalence under all three phenotype definitions, N=8{,}024.}}
  {\begin{tabular}{lrr}
  \toprule
  \bfseries Phenotype & \bfseries Positive & \bfseries Prevalence \\
  \midrule
  A (measured only) & 3{,}408 & 42.5\% \\
  B (primary) & 4{,}311 & 53.7\% \\
  C (broadest) & 4{,}555 & 56.8\% \\
  \bottomrule
  \end{tabular}}
\end{table}

Table~\ref{tab:varcount} summarizes the 45 raw NHANES variables retained
for phenotype construction, baseline predictors, and lifestyle-feature
derivation, grouped by role. Lifestyle variables were curated from 26 raw
questionnaire items into 11 derived features to avoid redundant
skip-pattern-laden columns; e.g., a Y/N activity question and its
conditional days/minutes follow-ups were collapsed into one weekly-minutes
total, with a structural zero (not a missing value) recorded when the
parent Y/N response was negative. The same structural-zero logic was
applied to alcohol and smoking status.

\textbf{Model predictors.} The baseline arm comprised age
(\texttt{RIDAGEYR}), sex (\texttt{is\_female}), race/ethnicity
(\texttt{RIDRETH3}), education (\texttt{DMDEDUC2}), family
income-to-poverty ratio (\texttt{INDFMPIR}), weight (\texttt{BMXWT}),
height (\texttt{BMXHT}), BMI (\texttt{BMXBMI}), and waist circumference
(\texttt{BMXWAIST}). The lifestyle-augmented arm additionally included
weekly vigorous activity (\texttt{total\_vigorous\_min\_wk}), weekly
moderate activity (\texttt{total\_moderate\_min\_wk}), daily sedentary
time (\texttt{sedentary\_min\_day}), drinking status
(\texttt{drinking\_status}), average drinks/day
(\texttt{alq130\_avg\_drinks\_filled}), smoking status
(\texttt{smoking\_status}), recent tobacco use
(\texttt{smoked\_recently}), weekday sleep duration (\texttt{SLD012}),
weekend sleep duration (\texttt{SLD013}), trouble sleeping
(\texttt{trouble\_sleeping}), and daytime sleepiness
(\texttt{SLQ120}).

Survey design variables (sample weights, primary sampling units, strata)
were retained in the processed dataset but were not used as model
predictors; all reported estimates describe predictive performance within
this analytic cohort and are not weighted to be nationally representative.

\begin{table*}[htbp]
\floatconts
  {tab:varcount}
  {\caption{NHANES variables by category, source modules, and role.}}
  {\begin{tabular}{lrl}
  \toprule
  \bfseries Category & \bfseries N vars & \bfseries Source modules \\
  \midrule
  Phenotype & 10 & P\_BPQ, P\_BPXO \\
  Demographic/SES & 5 & P\_DEMO \\
  Anthropometric & 4 & P\_BMX \\
  Lifestyle (raw) & 26 & P\_PAQ, P\_ALQ, P\_SLQ, P\_SMQ, P\_SMQRTU \\
  \bottomrule
  \end{tabular}}
\end{table*}

\textbf{Variable selection.} The lifestyle block was prespecified rather
than selected from the data. The study asks whether a block of commonly
recommended modifiable behaviors (physical activity, alcohol, smoking,
and sleep) adds value beyond a routine baseline; these domains follow the
guideline and meta-analytic evidence cited in Section~\ref{sec:intro}
\citep{whelton_2017_2018, barone_gibbs_physical_2021,
cecchini_alcohol_2024, hosseini_association_2024, virdis_cigarette_2010}.
Selecting features using the outcome on the same data would tune the
block to the data it is tested on and bias the incremental comparison.
Every variable in the nine NHANES modules used has a recorded inclusion
or exclusion decision with a stated reason; for example, snoring and
breathing-pause items were excluded as sleep-apnea screening rather than
lifestyle habits, age-restricted and interview-logistics items were
excluded, and raw bed and wake times were replaced by derived sleep
duration. Questionnaire skip patterns were respected, so that a ``no''
to a parent question became a structural zero rather than a missing
value. Variables were chosen from the study aims and the cited
literature without a formal clinical expert review. The full
variable-by-variable record, including candidates for future expansion
such as waist-to-hip ratio and diabetes status, is in the code repository.
 
\textbf{Cohort characteristics.} Table~\ref{tab:cohort_characteristics}
describes the 8{,}024 phenotype-eligible participants overall and by
hypertension status under Phenotypes B and A. These summaries are
unweighted and describe the analytic cohort rather than the U.S.\
population.
 
\begin{table*}[htbp]
\floatconts
  {tab:cohort_characteristics}
  {\caption{Characteristics of the analytic cohort, overall and by hypertension status under the primary Phenotype B (measured $\geq$130/80~mmHg or antihypertensive medication) and Phenotype A (measured BP only). Summaries are unweighted and describe the analytic cohort, not the U.S.\ population. Percentages use non-missing values; ``Missing'' counts missing values in the overall cohort. Age is top-coded at 80 and the income-to-poverty ratio at 5 in the public NHANES release. Daytime sleepiness is the share reporting feeling overly sleepy often or almost always.}}
  {\scriptsize\setlength{\tabcolsep}{2pt}\begin{tabular}{@{}lcccccc@{}}
  \toprule
  \bfseries Characteristic & \bfseries Overall & \bfseries B: yes & \bfseries B: no & \bfseries A: yes & \bfseries A: no & \bfseries Missing \\
   & $n = 8{,}024$ & $n = 4{,}311$ & $n = 3{,}713$ & $n = 3{,}408$ & $n = 4{,}616$ & \\
  \midrule
  \multicolumn{7}{@{}l}{\itshape Demographic/SES} \\
  Age, years, mean (SD) & 49.4 (18.3) & 57.7 (15.4) & 39.8 (16.6) & 56.2 (16.0) & 44.3 (18.2) & 0 \\
  Female, n (\%) & 4{,}049 (50.5) & 2{,}042 (47.4) & 2{,}007 (54.1) & 1{,}594 (46.8) & 2{,}455 (53.2) & 0 \\
  Race/ethnicity, n (\%) &  &  &  &  &  & 0 \\
  \quad Mexican American & 899 (11.2) & 375 (8.7) & 524 (14.1) & 309 (9.1) & 590 (12.8) &  \\
  \quad Other Hispanic & 810 (10.1) & 394 (9.1) & 416 (11.2) & 311 (9.1) & 499 (10.8) &  \\
  \quad Non-Hispanic White & 2{,}842 (35.4) & 1{,}540 (35.7) & 1{,}302 (35.1) & 1{,}166 (34.2) & 1{,}676 (36.3) &  \\
  \quad Non-Hispanic Black & 2{,}147 (26.8) & 1{,}362 (31.6) & 785 (21.1) & 1{,}103 (32.4) & 1{,}044 (22.6) &  \\
  \quad Non-Hispanic Asian & 928 (11.6) & 436 (10.1) & 492 (13.3) & 353 (10.4) & 575 (12.5) &  \\
  \quad Other/multiracial & 398 (5.0) & 204 (4.7) & 194 (5.2) & 166 (4.9) & 232 (5.0) &  \\
  Education, n (\%) &  &  &  &  &  & 381 \\
  \quad Less than 9th grade & 518 (6.8) & 324 (7.6) & 194 (5.8) & 263 (7.8) & 255 (6.0) &  \\
  \quad 9th--11th grade & 833 (10.9) & 505 (11.8) & 328 (9.8) & 411 (12.2) & 422 (9.9) &  \\
  \quad High school/GED & 1{,}858 (24.3) & 1{,}103 (25.8) & 755 (22.5) & 866 (25.6) & 992 (23.3) &  \\
  \quad Some college/AA & 2{,}533 (33.1) & 1{,}403 (32.8) & 1{,}130 (33.6) & 1{,}107 (32.8) & 1{,}426 (33.5) &  \\
  \quad College graduate+ & 1{,}901 (24.9) & 946 (22.1) & 955 (28.4) & 733 (21.7) & 1{,}168 (27.4) &  \\
  Income-to-poverty ratio, med. [IQR] & 2.19 [1.17, 4.17] & 2.24 [1.21, 4.15] & 2.15 [1.11, 4.22] & 2.16 [1.18, 4.02] & 2.22 [1.16, 4.35] & 1{,}041 \\
  \midrule
  \multicolumn{7}{@{}l}{\itshape Anthropometric} \\
  Weight, kg, mean (SD) & 83.7 (23.0) & 87.6 (23.6) & 79.1 (21.4) & 86.8 (23.3) & 81.4 (22.5) & 63 \\
  Height, cm, mean (SD) & 167.0 (10.0) & 167.1 (10.2) & 167.0 (9.8) & 167.2 (10.3) & 166.9 (9.9) & 71 \\
  BMI, kg/m$^2$, mean (SD) & 29.9 (7.5) & 31.2 (7.5) & 28.3 (7.1) & 30.9 (7.3) & 29.1 (7.5) & 79 \\
  Waist, cm, mean (SD) & 100.5 (17.4) & 105.3 (16.6) & 95.1 (16.6) & 104.4 (16.4) & 97.8 (17.5) & 244 \\
  \midrule
  \multicolumn{7}{@{}l}{\itshape Lifestyle} \\
  Vigorous activity, min/wk, med. [IQR] & 0 [0, 240] & 0 [0, 180] & 40 [0, 360] & 0 [0, 180] & 0 [0, 315] & 2 \\
  Moderate activity, min/wk, med. [IQR] & 180 [0, 690] & 150 [0, 600] & 225 [10, 840] & 150 [0, 600] & 180 [0, 720] & 1 \\
  Sedentary, min/day, med. [IQR] & 300 [180, 480] & 300 [180, 480] & 300 [180, 480] & 300 [180, 480] & 300 [180, 480] & 57 \\
  Drinking status, n (\%) &  &  &  &  &  & 297 \\
  \quad Current & 5{,}469 (70.8) & 2{,}758 (66.6) & 2{,}711 (75.6) & 2{,}210 (67.6) & 3{,}259 (73.1) &  \\
  \quad Former & 1{,}497 (19.4) & 1{,}006 (24.3) & 491 (13.7) & 740 (22.6) & 757 (17.0) &  \\
  \quad Never & 761 (9.8) & 375 (9.1) & 386 (10.8) & 321 (9.8) & 440 (9.9) &  \\
  Drinks/day, med. [IQR] & 1.0 [0.0, 2.0] & 1.0 [0.0, 2.0] & 1.0 [1.0, 3.0] & 1.0 [0.0, 2.0] & 1.0 [0.0, 2.0] & 307 \\
  Smoking status, n (\%) &  &  &  &  &  & 4 \\
  \quad Current & 1{,}432 (17.9) & 795 (18.5) & 637 (17.2) & 651 (19.1) & 781 (16.9) &  \\
  \quad Former & 1{,}818 (22.7) & 1{,}195 (27.7) & 623 (16.8) & 903 (26.5) & 915 (19.8) &  \\
  \quad Never & 4{,}770 (59.5) & 2{,}318 (53.8) & 2{,}452 (66.1) & 1{,}851 (54.4) & 2{,}919 (63.3) &  \\
  Recent tobacco use, n (\%) & 1{,}844 (23.9) & 975 (23.6) & 869 (24.2) & 796 (24.3) & 1{,}048 (23.5) & 295 \\
  Weekday sleep, h, mean (SD) & 7.6 (1.7) & 7.6 (1.7) & 7.6 (1.6) & 7.5 (1.7) & 7.6 (1.7) & 65 \\
  Weekend sleep, h, mean (SD) & 8.3 (1.8) & 8.1 (1.8) & 8.4 (1.7) & 8.1 (1.8) & 8.4 (1.8) & 66 \\
  Trouble sleeping, n (\%) & 2{,}290 (28.6) & 1{,}471 (34.1) & 819 (22.1) & 1{,}062 (31.2) & 1{,}228 (26.6) & 5 \\
  Daytime sleepiness, n (\%) & 2{,}089 (26.1) & 1{,}131 (26.3) & 958 (25.8) & 854 (25.1) & 1{,}235 (26.8) & 13 \\
  \bottomrule
  \end{tabular}}
\end{table*}

\section{Training and Hyperparameter Details}
\label{apd:training}

Each pipeline applied one-hot encoding to categorical features, median
imputation, standardization (Logistic Regression only), and SMOTE
\citep{chawla_smote_2002}, implemented via an \texttt{imbalanced-learn}
Pipeline \citep{lemaitre_imbalanced-learn_2016} so that all fitted steps
were refit within each training fold, never on validation or test data.
Models were implemented in scikit-learn \citep{pedregosa_scikit-learn_2012}:
Logistic Regression, Random Forest \citep{breiman_random_2001}, Gradient
Boosting \citep{friedman_greedy_2001}, and Bagged Decision Trees
\citep{breiman_bagging_1996}; XGBoost \citep{chen_xgboost_2016} used its
native implementation. Hyperparameters were selected via 5-fold stratified
cross-validation (ROC-AUC) on the fixed 90\% development set
\citep{kohavi_study_1995}: an exhaustive grid for Logistic Regression, and
25-candidate randomized search \citep{bergstra_random_2012} for the
remaining four models, run independently for each of the two feature
arms (Table~\ref{tab:searchspace}). Selected \textbf{classifier hyperparameters} (Table~\ref{tab:frozenparams})
were frozen and reused, unmodified, for every downstream analysis
(held-out evaluation, SHAP, both robustness checks) -- no hyperparameter
was re-selected using held-out test performance or any sensitivity-analysis
result. Two robustness checks deliberately varied other pipeline elements
while holding classifier hyperparameters fixed: the SMOTE sensitivity
analysis (Appendix~\ref{apd:robustness}) toggled SMOTE's presence, and the
Phenotype A analysis changed the prediction target.
Model-family comparisons are also sensitive to tuning: XGBoost required
substantial hyperparameter search to reach parity with the other four
models on this dataset size, so a comparison using library defaults
alone would have misleadingly suggested a real model-family difference
that tuning mostly closed.
\begin{table}[htbp]
\floatconts
  {tab:searchspace}
  {\caption{Hyperparameter search spaces. LR used an exhaustive grid; all
  others used 25-candidate randomized search.}}
  {\footnotesize\begin{tabular}{@{}lp{0.58\columnwidth}@{}}
  \toprule
  \bfseries Model & \bfseries Search space \\
  \midrule
  Logistic Regression & $C \in \{0.01, 0.1, 1, 10, 100\}$ \\
  \midrule
  Random Forest & \texttt{n\_estimators} $\in\{200,300,500\}$ \\
   & \texttt{max\_depth} $\in\{$None$,5,10,20\}$ \\
   & \texttt{min\_samples\_leaf} $\in\{1,2,5,10\}$ \\
   & \texttt{max\_features} $\in\{$sqrt, log2, 0.5$\}$ \\
  \midrule
  Gradient Boosting & \texttt{n\_estimators} $\in\{100,200,300\}$ \\
   & \texttt{learning\_rate} $\in\{.01,.05,.1,.2\}$ \\
   & \texttt{max\_depth} $\in\{2,3,4,5\}$ \\
   & \texttt{min\_samples\_leaf} $\in\{1,5,10,20\}$ \\
  \midrule
  Bagged Decision Tree & \texttt{n\_estimators} $\in\{200,300,500\}$ \\
   & base \texttt{max\_depth} $\in\{$None$,5,10,20\}$ \\
   & base \texttt{min\_samples\_leaf} $\in\{1,2,5,10\}$ \\
   & \texttt{max\_samples} $\in\{.5,.7,1.0\}$ \\
  \midrule
  XGBoost & \texttt{n\_estimators} $\in\{100,200,300,500\}$ \\
   & \texttt{learning\_rate} $\in\{.01,.05,.1,.2\}$ \\
   & \texttt{max\_depth} $\in\{3,4,5,6,8\}$ \\
   & \texttt{subsample} $\in\{.6,.8,1.0\}$ \\
   & \texttt{colsample\_bytree} $\in\{.6,.8,1.0\}$ \\
   & \texttt{min\_child\_weight} $\in\{1,3,5\}$ \\
  \bottomrule
  \end{tabular}}
\end{table}

\begin{table*}[htbp]
\floatconts
  {tab:frozenparams}
  {\caption{Selected hyperparameters, frozen for all downstream analyses.}}
  {\footnotesize\begin{tabular}{llp{9cm}}
  \toprule
  \bfseries Arm & \bfseries Model & \bfseries Selected hyperparameters \\
  \midrule
  Baseline & Logistic Regression & $C{=}10$ \\
  Baseline & Random Forest & n\_estimators=200, min\_samples\_leaf=10, max\_features=0.5, max\_depth=5 \\
  Baseline & Gradient Boosting & n\_estimators=100, min\_samples\_leaf=20, max\_depth=3, learning\_rate=0.05 \\
  Baseline & Bagged Decision Tree & n\_estimators=500, max\_samples=0.5, base min\_samples\_leaf=10, base max\_depth=5 \\
  Baseline & XGBoost & subsample=1.0, n\_estimators=300, min\_child\_weight=5, max\_depth=4, learning\_rate=0.01, colsample\_bytree=0.8 \\
  \midrule
  Lifestyle-augmented & Logistic Regression & $C{=}100$ \\
  Lifestyle-augmented & Random Forest & n\_estimators=500, min\_samples\_leaf=2, max\_features=0.5, max\_depth=5 \\
  Lifestyle-augmented & Gradient Boosting & n\_estimators=100, min\_samples\_leaf=10, max\_depth=4, learning\_rate=0.05 \\
  Lifestyle-augmented & Bagged Decision Tree & n\_estimators=300, max\_samples=0.5, base min\_samples\_leaf=10, base max\_depth=10 \\
  Lifestyle-augmented & XGBoost & subsample=1.0, n\_estimators=300, min\_child\_weight=5, max\_depth=4, learning\_rate=0.01, colsample\_bytree=0.8 \\
  \bottomrule
  \end{tabular}}
\end{table*}

\textbf{Search budget and sensitivity.} The 25-candidate budget for the
tree models was fixed before the search was run, applied uniformly
across the four tree-based model families so that tuning effort would
not differ between them, and not changed in response to results; random
search typically finds near-optimal configurations with modest budgets
when only a few hyperparameters matter \citep{bergstra_random_2012}.
Table~\ref{tab:tuning_summary} summarizes the full search log. In every
arm and model, several sampled configurations reached cross-validated
ROC-AUC within 0.002 of the best, and the margin between the best and
second-best configurations ($\leq$0.0005) was far smaller than the
cross-validation standard deviation of the selected configuration
(about 0.012). This indicates that the sampled candidates performed
similarly; it does not establish that the search converged, and modest
gains from a much larger search cannot be ruled out. As context only,
the tuned tree models reached the same cross-validated ROC-AUC as the
exhaustively tuned Logistic Regression (0.813 baseline, 0.815
lifestyle-augmented), and tuning substantially improved weak library
defaults (e.g., XGBoost 0.782 to 0.813 and Bagged Decision Trees 0.791
to 0.813 in the baseline arm).
 
\begin{table*}[htbp]
\floatconts
  {tab:tuning_summary}
  {\caption{Hyperparameter search summary, 5-fold development-set CV ROC-AUC. ``Default'' is library-default CV ROC-AUC before tuning; ``Best'' is the selected configuration; ``Gap to 2nd'' is its margin over the next-best candidate; ``Within 0.002'' counts candidates within 0.002 of the best; ``SD of best'' is the CV fold standard deviation of the selected configuration. Logistic Regression used an exhaustive five-value grid; tree models used 25 randomized candidates.}}
  {\footnotesize\setlength{\tabcolsep}{4pt}\begin{tabular}{@{}llcccccc@{}}
  \toprule
  \bfseries Arm & \bfseries Model & \bfseries Cand. & \bfseries Default & \bfseries Best & \bfseries Gap to 2nd & \bfseries Within 0.002 & \bfseries SD of best \\
  \midrule
  Baseline & Logistic Regression & 5 & 0.813 & 0.813 & 0.0000 & 5/5 & 0.013 \\
   & Random Forest & 25 & 0.796 & 0.812 & 0.0001 & 10/25 & 0.012 \\
   & Gradient Boosting & 25 & 0.810 & 0.812 & 0.0001 & 4/25 & 0.012 \\
   & Bagged Decision Tree & 25 & 0.791 & 0.813 & 0.0001 & 8/25 & 0.012 \\
   & XGBoost & 25 & 0.782 & 0.813 & 0.0003 & 5/25 & 0.012 \\
  \midrule
  Lifestyle-augmented & Logistic Regression & 5 & 0.815 & 0.815 & 0.0000 & 5/5 & 0.010 \\
   & Random Forest & 25 & 0.804 & 0.812 & 0.0000 & 11/25 & 0.013 \\
   & Gradient Boosting & 25 & 0.814 & 0.815 & 0.0005 & 5/25 & 0.013 \\
   & Bagged Decision Tree & 25 & 0.800 & 0.813 & 0.0004 & 8/25 & 0.012 \\
   & XGBoost & 25 & 0.785 & 0.815 & 0.0002 & 4/25 & 0.013 \\
  \bottomrule
  \end{tabular}}
\end{table*}

\section{Complete Predictive Results}
\label{apd:results}

Table~\ref{tab:full_cv} reports 5-fold development CV performance (mean
$\pm$ SD) for the frozen configurations above. Table~\ref{tab:full_test}
reports held-out test performance with bootstrap 95\% confidence
intervals \citep{efron_introduction_1993}. Table~\ref{tab:full_paired}
reports the paired lifestyle-minus-baseline delta per model (main-text
Figure~\ref{fig:test_delta} visualizes the ROC-AUC column of this table).

\begin{table*}[htbp]
\floatconts
  {tab:full_cv}
  {\caption{Development-set 5-fold CV performance, frozen hyperparameters.}}
  {\small\begin{tabular}{llcccc}
  \toprule
  \bfseries Arm & \bfseries Model & \bfseries ROC-AUC & \bfseries SD & \bfseries PR-AUC & \bfseries SD \\
  \midrule
  Baseline & Logistic Regression & 0.8128 & 0.0125 & 0.8102 & 0.0171 \\
  Baseline & Random Forest & 0.8122 & 0.0123 & 0.8113 & 0.0126 \\
  Baseline & Gradient Boosting & 0.8118 & 0.0120 & 0.8097 & 0.0127 \\
  Baseline & Bagged Decision Tree & 0.8125 & 0.0122 & 0.8114 & 0.0122 \\
  Baseline & XGBoost & 0.8130 & 0.0124 & 0.8125 & 0.0131 \\
  Lifestyle-augmented & Logistic Regression & 0.8154 & 0.0100 & 0.8135 & 0.0093 \\
  Lifestyle-augmented & Random Forest & 0.8117 & 0.0128 & 0.8102 & 0.0134 \\
  Lifestyle-augmented & Gradient Boosting & 0.8153 & 0.0130 & 0.8114 & 0.0145 \\
  Lifestyle-augmented & Bagged Decision Tree & 0.8131 & 0.0122 & 0.8078 & 0.0130 \\
  Lifestyle-augmented & XGBoost & 0.8150 & 0.0128 & 0.8116 & 0.0140 \\
  \bottomrule
  \end{tabular}}
\end{table*}

\begin{table*}[htbp]
\floatconts
  {tab:full_test}
  {\caption{Held-out test performance with bootstrap 95\% confidence intervals.}}
  {\small\begin{tabular}{llccc}
  \toprule
  \bfseries Arm & \bfseries Model & \bfseries ROC-AUC [95\% CI] & \bfseries PR-AUC [95\% CI] \\
  \midrule
  Baseline & Logistic Regression & 0.809 [0.778, 0.838] & 0.798 [0.755, 0.841] \\
  Baseline & Random Forest & 0.804 [0.773, 0.834] & 0.792 [0.747, 0.837] \\
  Baseline & Gradient Boosting & 0.809 [0.779, 0.839] & 0.802 [0.760, 0.844] \\
  Baseline & Bagged Decision Tree & 0.807 [0.776, 0.837] & 0.796 [0.753, 0.838] \\
  Baseline & XGBoost & 0.810 [0.779, 0.840] & 0.796 [0.754, 0.840] \\
  Lifestyle-augmented & Logistic Regression & 0.807 [0.777, 0.837] & 0.791 [0.748, 0.834] \\
  Lifestyle-augmented & Random Forest & 0.804 [0.772, 0.832] & 0.791 [0.746, 0.835] \\
  Lifestyle-augmented & Gradient Boosting & 0.808 [0.779, 0.838] & 0.798 [0.756, 0.840] \\
  Lifestyle-augmented & Bagged Decision Tree & 0.807 [0.778, 0.838] & 0.794 [0.751, 0.838] \\
  Lifestyle-augmented & XGBoost & 0.807 [0.777, 0.837] & 0.795 [0.754, 0.838] \\
  \bottomrule
  \end{tabular}}
\end{table*}

\begin{table*}[htbp]
\floatconts
  {tab:full_paired}
  {\caption{Paired bootstrap deltas (lifestyle-augmented minus baseline), held-out test.}}
  {\footnotesize\begin{tabular}{lcc}
  \toprule
  \bfseries Model & \bfseries $\Delta$ ROC-AUC [95\% CI] & \bfseries $\Delta$ PR-AUC [95\% CI] \\
  \midrule
  Logistic Regression & $-$0.0018 [$-$0.0083, 0.0052] & $-$0.0072 [$-$0.0187, 0.0060] \\
  Random Forest & $-$0.0009 [$-$0.0044, 0.0028] & $-$0.0010 [$-$0.0070, 0.0052] \\
  Gradient Boosting & $-$0.0018 [$-$0.0085, 0.0054] & $-$0.0044 [$-$0.0159, 0.0077] \\
  Bagged Decision Tree & 0.0002 [$-$0.0058, 0.0061] & $-$0.0019 [$-$0.0117, 0.0081] \\
  XGBoost & $-$0.0028 [$-$0.0075, 0.0026] & $-$0.0007 [$-$0.0080, 0.0077] \\
  \bottomrule
  \end{tabular}}
\end{table*}

\section{Interpretability Details}
\label{apd:interpretability}
SHAP values \citep{lundberg_unified_2017} were computed with
\texttt{TreeExplainer} \citep{lundberg_local_2020} for Random Forest,
Gradient Boosting, and XGBoost. \texttt{TreeExplainer} does not support
\texttt{BaggingClassifier}; a model-agnostic permutation explainer was
substituted for that model only. Cross-model agreement
(Table~\ref{tab:shap_stability}) was assessed over the post-encoding
feature space (19 baseline, 36 lifestyle-augmented columns) and, for the
matched comparison, over the 19 baseline columns shared by both arms
(Appendix~\ref{apd:shapcontrols}). The illustrative top-feature visualization (Figure~\ref{fig:top_features})
instead aggregates one-hot-encoded categorical predictors (e.g.,
race/ethnicity) back to their source variable, for interpretability: SHAP
values for a predictor's encoded columns were summed within each sample
before absolute values were averaged across samples. These are two
distinct units of analysis and are not directly comparable. Cross-model agreement was quantified two ways,
following \citet{krishna_disagreement_2022}'s characterization of
explanation disagreement across models: top-10 feature-membership
Jaccard overlap, and Spearman rank correlation over the complete
importance ordering (Table~\ref{tab:shap_stability}). $\rho$ magnitudes
are reported descriptively; p-values are not treated as a formal
hypothesis test given the small, non-independent feature set.

\begin{figure}[htbp]
\floatconts
  {fig:top_features}
  {\caption{Top-10 predictor-level SHAP importance for Logistic Regression,
  which achieved the highest development-set CV ROC-AUC in the
  lifestyle-augmented arm.}}
  {\includegraphics[width=\linewidth]
  {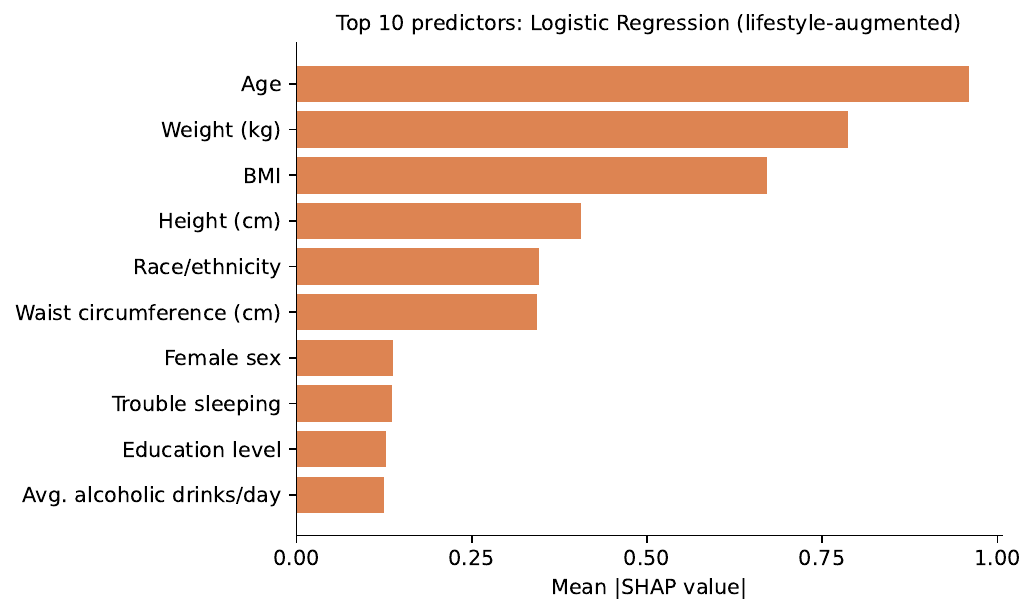}}
\end{figure}

\section{Robustness and Sensitivity Analyses}
\label{apd:robustness}

\textbf{Phenotype A.} All ten frozen hyperparameter configurations were
refit under Phenotype A (Table~\ref{tab:pheno_a_full}), using the
identical 90/10 participant split as the primary analysis (stratified on
Phenotype B) so that only the label, not cohort composition, varied. The
split is not meaningfully imbalanced for Phenotype A: its prevalence is
42.6\% in the development set (3{,}074/7{,}221) and 41.6\% in the test
set (334/803), against 42.5\% overall.

\begin{table}[htbp]
\floatconts
  {tab:pheno_a_full}
  {\caption{Phenotype A paired deltas (lifestyle-augmented minus baseline), held-out test.}}
  {\footnotesize\begin{tabular}{lc}
  \toprule
  \bfseries Model & \bfseries $\Delta$ ROC-AUC [95\% CI] \\
  \midrule
  Logistic Regression & 0.0009 [$-$0.0090, 0.0108] \\
  Random Forest & 0.0021 [$-$0.0052, 0.0088] \\
  Gradient Boosting & $-$0.0006 [$-$0.0115, 0.0093] \\
  Bagged Decision Tree & 0.0086 [$-$0.0020, 0.0199] \\
  XGBoost & 0.0038 [$-$0.0047, 0.0120] \\
  \bottomrule
  \end{tabular}}
\end{table}

\textbf{SMOTE sensitivity.} Using the same frozen hyperparameters, each
model was refit with SMOTE included and excluded (all other pipeline
steps identical). Table~\ref{tab:smote_full} reports the 10 paired
SMOTE-minus-no-SMOTE comparisons across five models and two feature arms.
All 10 paired-bootstrap 95\% AUROC confidence intervals included zero.

\begin{table}[htbp]
\floatconts
  {tab:smote_full}
  {\caption{SMOTE-minus-no-SMOTE paired deltas with 95\% CIs, held-out
  test, frozen hyperparameters. ``Lifestyle'' is the lifestyle-augmented
  arm.}}
  {\scriptsize\setlength{\tabcolsep}{4pt}\begin{tabular}{@{}llc@{}}
  \toprule
  \bfseries Arm & \bfseries Model & \bfseries $\Delta$ ROC-AUC [95\% CI] \\
  \midrule
  Baseline & Logistic Regression & 0.0002 [$-$0.0010, 0.0013] \\
   & Random Forest & 0.0029 [$-$0.0005, 0.0063] \\
   & Gradient Boosting & 0.0006 [$-$0.0024, 0.0036] \\
   & Bagged Decision Tree & 0.0012 [$-$0.0016, 0.0041] \\
   & XGBoost & 0.0005 [$-$0.0024, 0.0032] \\
  \midrule
  Lifestyle & Logistic Regression & $-$0.0006 [$-$0.0018, 0.0008] \\
   & Random Forest & 0.0021 [$-$0.0012, 0.0058] \\
   & Gradient Boosting & $-$0.0023 [$-$0.0069, 0.0020] \\
   & Bagged Decision Tree & $-$0.0006 [$-$0.0048, 0.0035] \\
   & XGBoost & $-$0.0031 [$-$0.0064, 0.0001] \\
  \bottomrule
  \end{tabular}}
\end{table}

\section{SHAP Agreement Controls}
\label{apd:shapcontrols}
 
These controls test whether the lower full-column SHAP agreement in the
lifestyle-augmented arm (Table~\ref{tab:shap_stability}) reflects the
lifestyle features themselves or the larger number of columns being
ranked (36 vs.\ 19). The matched-space, predictor-level, and
domain-level views reuse the SHAP values already computed for the same
300 test participants, so no model is refit; agreement is the mean over
the ten model pairs, and paired bootstrap intervals resample the
explained participants and therefore exclude model-refit variability.
 
\textbf{Matched comparison space.} The lifestyle arm's agreement is
recomputed over only the 19 baseline columns that the baseline arm
ranks. Table~\ref{tab:shap_robustness} reports this comparison for the
primary phenotype, Phenotype A, and the no-SMOTE refit. In all three
settings, full-column Spearman agreement declined, while agreement over
the shared 19 columns did not decline. Under Phenotype A the
shared-column interval lay above zero ($+$0.029 [$+$0.008, $+$0.053]).
 
\textbf{Shuffled-lifestyle control.} Each lifestyle predictor was
permuted independently within the development and test sets, preserving
the 36-column feature space and each predictor's marginal distribution
while disrupting its observed associations with other predictors and
with the outcome. All five models were refit with the frozen
hyperparameters and SHAP was recomputed. This is a single permutation
realization, and independent permutation can create unrealistic
combinations of predictor values, so it supports but does not prove the
interpretation in Section~\ref{sec:results}. Shuffled columns received
5.3\% of total mean $|$SHAP$|$, compared with 11.3\% for the real
lifestyle columns.
 
\textbf{Coarser units.} When one-hot columns are summed back to source
predictors, mean pairwise Spearman agreement is 0.884 in the lifestyle
arm (20 predictors) vs.\ 0.770 in the baseline arm (9 predictors); the
lists differ in length, so this comparison is descriptive only. At the
domain level (Table~\ref{tab:shap_domain}), all five models rank
demographic/SES first and anthropometric second, and the four lifestyle
domains together account for about 13\% of net grouped SHAP magnitude.
Disagreement is concentrated among the small lifestyle domains, whose
order varies across models. Under Phenotype A, domain-level agreement
was 0.84, with the same top two domains.
 
\begin{table*}[htbp]
\floatconts
  {tab:shap_robustness}
  {\caption{Cross-model SHAP agreement by setting (mean pairwise over five models; Base.\ = baseline arm, Life.\ = lifestyle-augmented arm). ``All'' uses each arm's full encoded column set (19 vs.\ 36); ``shared'' restricts the lifestyle arm to the 19 baseline columns, so the baseline value is unchanged. Jaccard is top-10 overlap. $\Delta$ is lifestyle minus baseline; 95\% CIs are paired bootstraps over the 300 explained participants and do not include model-refit variability. Chance-level top-10 Jaccard is 0.365 for 19 columns and 0.167 for 36.}}
  {\scriptsize\setlength{\tabcolsep}{4pt}\begin{tabular}{llccccc}
  \toprule
  \bfseries Setting & \bfseries Metric & \bfseries Base. & \bfseries Life., all & \bfseries $\Delta$ all [95\% CI] & \bfseries Life., shared & \bfseries $\Delta$ shared [95\% CI] \\
  \midrule
  Primary & Spearman $\rho$ & 0.873 & 0.806 & $-$0.067 [$-$0.081, $-$0.049] & 0.863 & $-$0.010 [$-$0.021, $+$0.010] \\
   & Jaccard & 0.821 & 0.699 & $-$0.122 [$-$0.189, $-$0.045] & 0.761 & $-$0.061 [$-$0.104, $+$0.030] \\
  \midrule
  Phenotype A & Spearman $\rho$ & 0.830 & 0.794 & $-$0.036 [$-$0.053, $-$0.013] & 0.858 & $+$0.029 [$+$0.008, $+$0.053] \\
   & Jaccard & 0.686 & 0.668 & $-$0.018 [$-$0.074, $-$0.007] & 0.742 & $+$0.056 [0.000, $+$0.126] \\
  \midrule
  No SMOTE & Spearman $\rho$ & 0.864 & 0.795 & $-$0.069 [$-$0.088, $-$0.059] & 0.873 & $+$0.008 [$-$0.017, $+$0.017] \\
   & Jaccard & 0.773 & 0.702 & $-$0.070 [$-$0.099, $-$0.003] & 0.758 & $-$0.015 [$-$0.018, $+$0.030] \\
  \bottomrule
  \end{tabular}}
\end{table*}
 
\begin{table}[htbp]
\floatconts
  {tab:shap_domain}
  {\caption{Domain-level SHAP importance, lifestyle-augmented arm, primary phenotype. Per-sample signed SHAP values are summed within each domain before absolute values are averaged over samples (the net grouped contribution, as in Figure~\ref{fig:top_features}); opposing attributions within a domain can cancel. ``Mean share'' averages each domain's share of net grouped SHAP magnitude across the five models; remaining columns give each model's domain rank. Mean pairwise Spearman agreement over the six domains is 0.903 [$+$0.874, $+$0.931].}}
  {\scriptsize\setlength{\tabcolsep}{3pt}\begin{tabular}{@{}lcccccc@{}}
  \toprule
  \bfseries Domain & \bfseries Share & \bfseries LR & \bfseries RF & \bfseries GB & \bfseries BDT & \bfseries XGB \\
  \midrule
  Demographic/SES & 63.1\% & 1 & 1 & 1 & 1 & 1 \\
  Anthropometric & 24.3\% & 2 & 2 & 2 & 2 & 2 \\
  Sleep & 4.9\% & 3 & 4 & 3 & 3 & 3 \\
  Smoking & 3.0\% & 5 & 3 & 4 & 4 & 4 \\
  Alcohol & 2.4\% & 4 & 6 & 6 & 6 & 6 \\
  Activity & 2.3\% & 6 & 5 & 5 & 5 & 5 \\
  \bottomrule
  \end{tabular}}
\end{table}
 
\section{Calibration and Decision-Curve Analysis}
\label{apd:calibration}
 
The frozen models were evaluated on the same held-out test set as the
main analysis (Table~\ref{tab:calibration};
Figure~\ref{fig:calibration}). Brier-score differences were negligible,
and all five paired intervals included zero. Both arms slightly
underestimate risk on average: mean predicted risk was 0.50--0.52
against an observed rate of 0.537, consistent with the positive
calibration-in-the-large values. Separately, calibration slopes at or
somewhat above 1 indicate that predictions are, if anything, slightly
insufficiently spread out rather than overconfident. Under the primary
phenotype the negatives are the minority class, so SMOTE oversamples
them during training; this may contribute to the downward shift seen in
both arms.
 
An exploratory decision-curve analysis \citep{vickers_decision_2006}
compared net benefit across risk thresholds from 0.05 to 0.80
(Figure~\ref{fig:decision_curve}). The two arms' curves lie close
together. Of the 20 paired comparisons at four thresholds
(Table~\ref{tab:net_benefit}), 19 intervals included zero; the
exception, Bagged Decision Trees at 0.5 ($+$0.020 [$+$0.005, $+$0.036]),
is isolated among correlated, unadjusted comparisons, and the other four
models' differences at that threshold differ in sign. These thresholds
are not clinically validated decision points for this cross-sectional
classification task, so the curves are not interpreted as evidence of
deployable clinical utility for either arm.
 
\begin{table*}[htbp]
\floatconts
  {tab:calibration}
  {\caption{Calibration on the held-out test set, frozen models (baseline / lifestyle-augmented). Calibration-in-the-large is the intercept with the slope fixed at 1; calibration slope is estimated separately with a free intercept. Both are point estimates. Observed prevalence is 0.537. $\Delta$ Brier is lifestyle minus baseline with a paired bootstrap 95\% CI (lower Brier is better).}}
  {\footnotesize\setlength{\tabcolsep}{4pt}\begin{tabular}{lccccc}
  \toprule
  \bfseries Model & \bfseries Brier & \bfseries Cal.\ slope & \bfseries CITL & \bfseries Mean pred. & \bfseries $\Delta$ Brier [95\% CI] \\
  \midrule
  Logistic Regression & 0.178 / 0.178 & 1.01 / 0.99 & $+$0.19 / $+$0.18 & 0.503 / 0.504 & 0.0000 [$-$0.0031, $+$0.0030] \\
  Random Forest & 0.179 / 0.179 & 1.13 / 1.10 & $+$0.17 / $+$0.16 & 0.505 / 0.506 & $-$0.0001 [$-$0.0016, $+$0.0013] \\
  Gradient Boosting & 0.176 / 0.176 & 1.11 / 1.05 & $+$0.13 / $+$0.09 & 0.513 / 0.520 & $-$0.0003 [$-$0.0034, $+$0.0028] \\
  Bagged Decision Tree & 0.178 / 0.177 & 1.05 / 1.00 & $+$0.17 / $+$0.15 & 0.505 / 0.510 & $-$0.0009 [$-$0.0036, $+$0.0018] \\
  XGBoost & 0.177 / 0.177 & 1.21 / 1.19 & $+$0.14 / $+$0.13 & 0.511 / 0.513 & $+$0.0005 [$-$0.0017, $+$0.0024] \\
  \bottomrule
  \end{tabular}}
\end{table*}
 
\begin{table*}[htbp]
\floatconts
  {tab:net_benefit}
  {\caption{Exploratory decision-curve analysis: paired lifestyle-minus-baseline differences in net benefit at four risk thresholds $p_t$, held-out test set, with paired bootstrap 95\% CIs. These 20 comparisons are correlated and unadjusted; the single interval excluding zero is in bold. Thresholds are illustrative, not clinically validated decision points.}}
  {\scriptsize\setlength{\tabcolsep}{3pt}\begin{tabular}{lcccc}
  \toprule
  \bfseries Model & \bfseries $p_t = 0.2$ & \bfseries $p_t = 0.3$ & \bfseries $p_t = 0.4$ & \bfseries $p_t = 0.5$ \\
  \midrule
  Logistic Regression & $+$0.002 [$-$0.005, $+$0.009] & $-$0.003 [$-$0.014, $+$0.009] & $-$0.002 [$-$0.014, $+$0.011] & $+$0.006 [$-$0.006, $+$0.020] \\
  Random Forest & $+$0.001 [$-$0.006, $+$0.007] & $+$0.002 [$-$0.003, $+$0.007] & $+$0.006 [$-$0.0004, $+$0.013] & $+$0.006 [$-$0.004, $+$0.017] \\
  Gradient Boosting & $-$0.002 [$-$0.007, $+$0.003] & $-$0.009 [$-$0.019, $+$0.001] & $+$0.001 [$-$0.012, $+$0.012] & $-$0.002 [$-$0.016, $+$0.011] \\
  Bagged Decision Tree & $-$0.004 [$-$0.010, $+$0.002] & $+$0.002 [$-$0.008, $+$0.012] & $+$0.003 [$-$0.008, $+$0.015] & \textbf{$+$0.020 [$+$0.005, $+$0.036]} \\
  XGBoost & $+$0.001 [$-$0.003, $+$0.005] & $-$0.006 [$-$0.015, $+$0.002] & $+$0.003 [$-$0.007, $+$0.015] & $-$0.004 [$-$0.019, $+$0.011] \\
  \bottomrule
  \end{tabular}}
\end{table*}
 
\begin{figure*}[htbp]
\floatconts
  {fig:calibration}
  {\caption{Reliability diagrams on the held-out test set, baseline vs.\
  lifestyle-augmented arm, frozen models. The dashed line is perfect
  calibration.}}
  {\includegraphics[width=\textwidth]{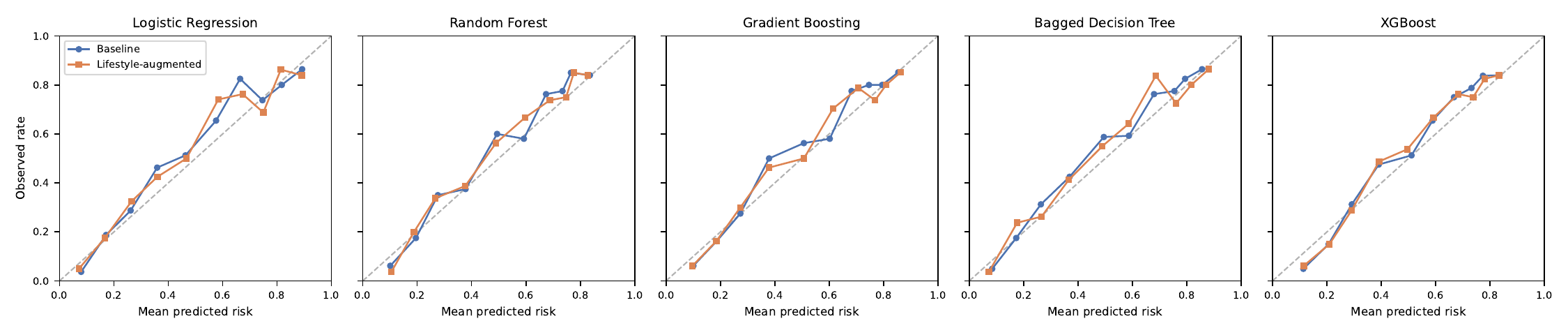}}
\end{figure*}
 
\begin{figure*}[htbp]
\floatconts
  {fig:decision_curve}
  {\caption{Exploratory decision curves on the held-out test set,
  baseline vs.\ lifestyle-augmented arm, with treat-all and treat-none
  references. Thresholds are illustrative, not validated decision
  points.}}
  {\includegraphics[width=\textwidth]{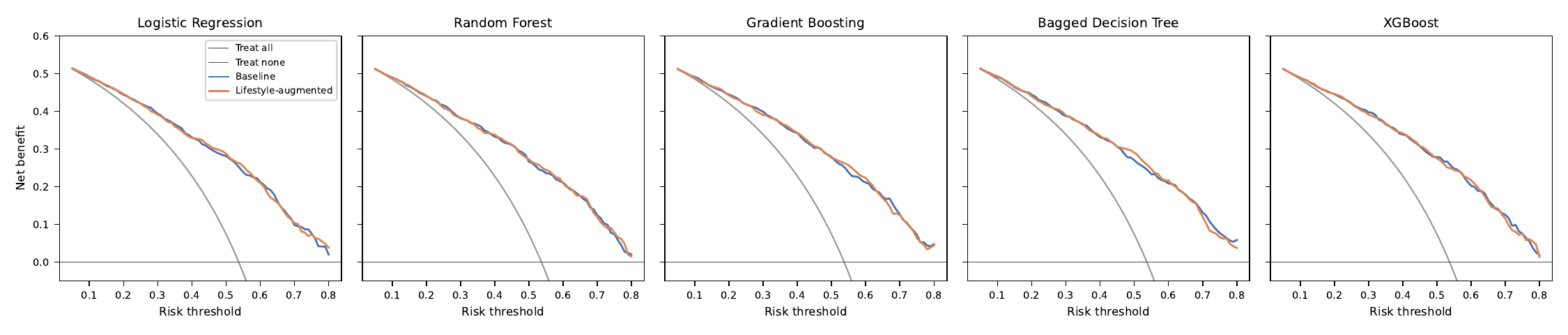}}
\end{figure*}
 
\section{Lifestyle--Baseline Redundancy Diagnostics}
\label{apd:redundancy}
 
These diagnostics use the development set only; the test set was not
used. Each statistic uses participants with non-missing values for the
variables involved, so sample sizes differ slightly across variables.
Pairwise Spearman correlations between lifestyle variables and baseline
predictors were at most moderate (Figure~\ref{fig:redundancy}): the
largest involved age (vigorous activity $-$0.30; drinks/day $-$0.28;
former smoking and former drinking $+$0.26 each), and against BMI or
waist circumference every $|\rho| \leq 0.17$. When all baseline
predictors were used jointly to predict each lifestyle variable,
cross-validated $R^2$ was at most 0.11 for continuous measures and
ROC-AUC was 0.63--0.76 for binary indicators (Table~\ref{tab:redundancy}).
 
To describe overlap with the outcome, each lifestyle variable was also
correlated with the outcome residual, defined as Phenotype B minus the
out-of-fold probability from a baseline-only model. Several
associations weakened substantially, for example vigorous activity
($-$0.16 to $-$0.05), former drinking ($+$0.13 to $+$0.02), former
smoking ($+$0.13 to $+$0.01), and moderate activity ($-$0.10 to
$-$0.01); trouble sleeping was a partial exception ($+$0.14 to
$+$0.08). This is consistent with overlapping predictive information
between the lifestyle block and the full baseline, but the analysis is
descriptive and does not establish conditional independence or a causal
explanation.
 
\begin{table*}[htbp]
\floatconts
  {tab:redundancy}
  {\caption{Lifestyle--baseline redundancy diagnostics, development set only. ``$\rho$ with outcome'' is the Spearman correlation with Phenotype B; ``$\rho$ with residual'' is the correlation with the outcome minus the out-of-fold probability from a baseline-only model. ``Predicted from baseline'' is the 5-fold cross-validated performance (mean $\pm$ SD) of predicting the lifestyle variable from all baseline predictors ($R^2$ for continuous, ROC-AUC for binary indicators). Each statistic uses participants with non-missing values. These are descriptive and do not establish conditional independence or causation.}}
  {\footnotesize\setlength{\tabcolsep}{4pt}\begin{tabular}{lccc}
  \toprule
  \bfseries Lifestyle variable & \bfseries $\rho$ with outcome & \bfseries $\rho$ with residual & \bfseries Predicted from baseline \\
  \midrule
  Vigorous activity (min/wk) & $-$0.16 & $-$0.05 & $R^2$ 0.08 $\pm$ 0.03 \\
  Trouble sleeping & $+$0.14 & $+$0.08 & AUC 0.63 $\pm$ 0.02 \\
  Drinking status: former & $+$0.13 & $+$0.02 & AUC 0.68 $\pm$ 0.02 \\
  Smoking status: never & $-$0.13 & $-$0.04 & AUC 0.73 $\pm$ 0.01 \\
  Smoking status: former & $+$0.13 & $+$0.01 & AUC 0.71 $\pm$ 0.01 \\
  Drinking status: current & $-$0.10 & $-$0.01 & AUC 0.69 $\pm$ 0.01 \\
  Moderate activity (min/wk) & $-$0.10 & $-$0.01 & $R^2$ 0.04 $\pm$ 0.02 \\
  Average drinks/day & $-$0.10 & $+$0.01 & $R^2$ 0.11 $\pm$ 0.02 \\
  Weekend sleep duration & $-$0.09 & $-$0.02 & $R^2$ $-$0.01 $\pm$ 0.01 \\
  Sedentary time (min/day) & $+$0.04 & $+$0.01 & $R^2$ 0.07 $\pm$ 0.02 \\
  Drinking status: never & $-$0.03 & $-$0.01 & AUC 0.70 $\pm$ 0.02 \\
  Smoking status: current & $+$0.02 & $+$0.04 & AUC 0.76 $\pm$ 0.02 \\
  Weekday sleep duration & $-$0.02 & 0.00 & $R^2$ $-$0.02 $\pm$ 0.02 \\
  Recent tobacco use & $-$0.01 & $+$0.03 & AUC 0.75 $\pm$ 0.01 \\
  Daytime sleepiness & 0.00 & $+$0.01 & $R^2$ 0.02 $\pm$ 0.02 \\
  \bottomrule
  \end{tabular}}
\end{table*}
 
\begin{figure}[htbp]
\floatconts
  {fig:redundancy}
  {\caption{Pairwise Spearman correlations between lifestyle variables
  and numeric or binary baseline predictors, development set.
  Race/ethnicity and education indicators are omitted from the figure
  but included in Table~\ref{tab:redundancy}'s joint prediction.}}
  {\includegraphics[width=\linewidth]{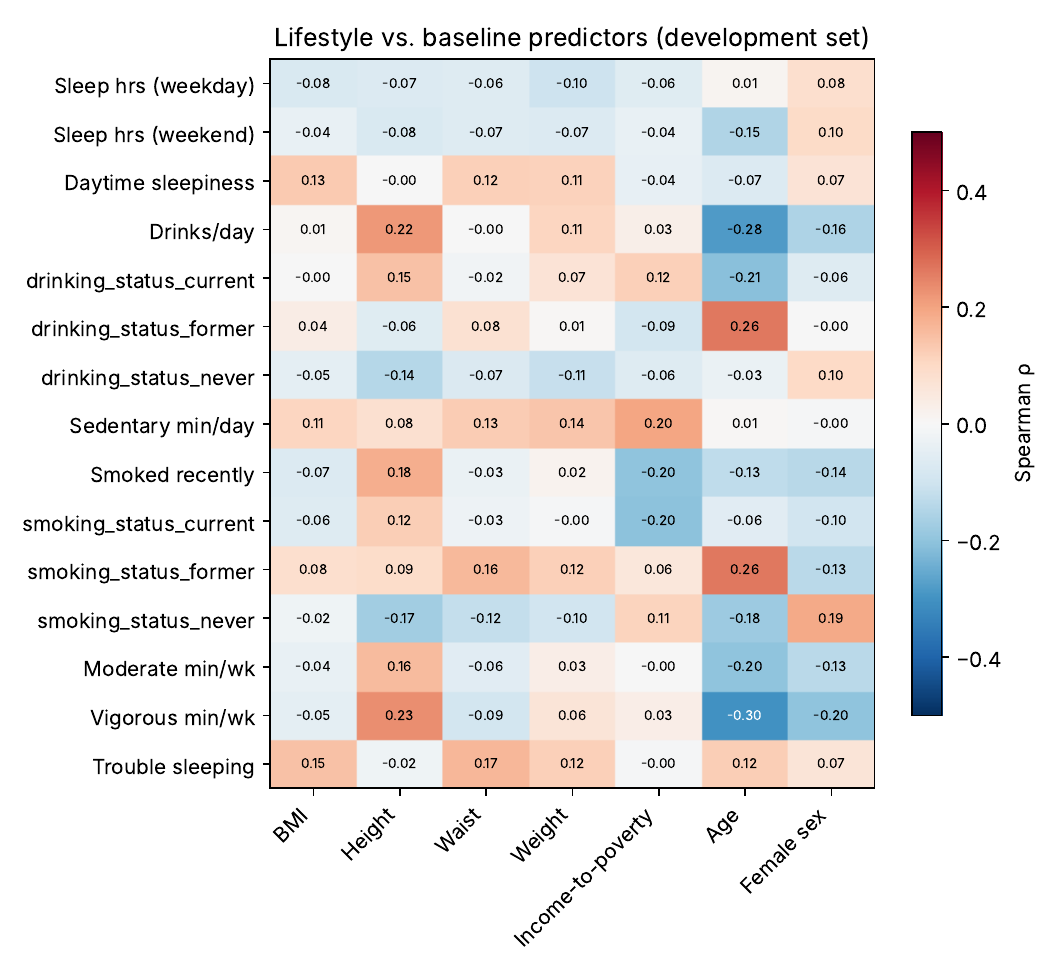}}
\end{figure}
 
\section{Untreated-Adult Subgroup}
\label{apd:untreated}
 
This exploratory analysis was restricted to participants definitely not
taking antihypertensive medication according to the questionnaire skip
pattern. The 14 participants with unknown treatment status were
excluded rather than assumed untreated, leaving $n = 5{,}697$
(development 5{,}131; test 566). Because the medication clause of
Phenotype B no longer applies, the outcome is measured-BP Phenotype A
(prevalence about 34\%). Hyperparameters are the frozen values tuned
for the primary phenotype in the full cohort and were not retuned.
 
All five point estimates favored lifestyle augmentation
(Table~\ref{tab:untreated}). Unadjusted intervals excluded zero for
Bagged Decision Trees and, narrowly, XGBoost (lower bound $+$0.00001);
the other three included zero. The subgroup test set is small, and the
five comparisons are not adjusted for multiplicity. Because the analysis
changes both the population and the outcome definition, the pattern
may reflect differences in treatment status, in the outcome definition,
or in the characteristics of the restricted population, and this
analysis cannot distinguish among them. It is a sensitivity analysis,
not an estimate of a treatment effect.
 
\begin{table*}[htbp]
\floatconts
  {tab:untreated}
  {\caption{Exploratory untreated-adult subgroup: participants definitely not taking antihypertensive medication, measured-BP Phenotype A outcome (development $n = 5{,}131$, prevalence 35.0\%; test $n = 566$, prevalence 34.3\%). Hyperparameters are the frozen primary-analysis values, not retuned. $\Delta$ is lifestyle minus baseline with a paired bootstrap 95\% CI; intervals excluding zero are in bold. The five comparisons are not adjusted for multiplicity.}}
  {\footnotesize\setlength{\tabcolsep}{4pt}\begin{tabular}{lccc}
  \toprule
  \bfseries Model & \bfseries Baseline AUC [95\% CI] & \bfseries Lifestyle AUC [95\% CI] & \bfseries $\Delta$ [95\% CI] \\
  \midrule
  Logistic Regression & 0.717 [0.675, 0.756] & 0.718 [0.673, 0.758] & $+$0.0003 [$-$0.0142, $+$0.0125] \\
  Random Forest & 0.702 [0.657, 0.748] & 0.710 [0.665, 0.753] & $+$0.0078 [$-$0.0068, $+$0.0216] \\
  Gradient Boosting & 0.714 [0.669, 0.756] & 0.720 [0.677, 0.763] & $+$0.0060 [$-$0.0091, $+$0.0221] \\
  Bagged Decision Tree & 0.700 [0.655, 0.746] & 0.719 [0.675, 0.760] & \textbf{$+$0.0181 [$+$0.0020, $+$0.0348]} \\
  XGBoost & 0.706 [0.660, 0.750] & 0.718 [0.674, 0.760] & \textbf{$+$0.0119 [$+$0.00001, $+$0.0241]} \\
  \bottomrule
  \end{tabular}}
\end{table*}

\section{Feature-Subset Ablations}
\label{apd:ablation}
 
Table~\ref{tab:ablation} reports held-out ROC-AUC on the original test
set for nested and single-block predictor subsets. Each subset reuses
the frozen hyperparameters of the corresponding arm and was not
retuned, so the smaller subsets may be slightly disadvantaged. Age alone
recovers most of the baseline's discrimination, and age plus BMI
recovers nearly all of it. Adding the anthropometric block to
demographic/SES predictors raised ROC-AUC by $+$0.009 to $+$0.014 across
models; the paired interval excluded zero for Logistic Regression
($+$0.014 [$+$0.0001, $+$0.028]) and included zero for the other four.
Lifestyle predictors alone discriminated above chance (0.629--0.648)
but less well than age alone, which is not by itself evidence of
clinical usefulness. Adding lifestyle to the full baseline showed no
consistent detectable improvement in held-out ROC-AUC ($-$0.003 to
0.000).
 
\begin{table*}[htbp]
\floatconts
  {tab:ablation}
  {\caption{Feature-subset ablations: held-out ROC-AUC on the original test set by predictor subset ($k$ = number of source predictors) and model. Subsets reuse the frozen hyperparameters of the corresponding arm (baseline arm for baseline subsets, lifestyle-augmented arm for subsets including lifestyle) and were not retuned, so they may be slightly disadvantaged. The last column gives the range across the five models of the paired difference from the full baseline.}}
  {\footnotesize\setlength{\tabcolsep}{4pt}\begin{tabular}{lccccccc}
  \toprule
  \bfseries Predictors & \bfseries $k$ & \bfseries LR & \bfseries RF & \bfseries GB & \bfseries BDT & \bfseries XGB & \bfseries $\Delta$ vs.\ full baseline \\
  \midrule
  Age only & 1 & 0.780 & 0.775 & 0.776 & 0.775 & 0.775 & $-$0.035 to $-$0.029 \\
  Age + BMI & 2 & 0.796 & 0.795 & 0.793 & 0.797 & 0.792 & $-$0.018 to $-$0.010 \\
  Demographic/SES only & 5 & 0.795 & 0.793 & 0.800 & 0.798 & 0.801 & $-$0.014 to $-$0.009 \\
  Anthropometric only & 4 & 0.681 & 0.685 & 0.686 & 0.686 & 0.685 & $-$0.127 to $-$0.119 \\
  Full baseline & 9 & 0.809 & 0.804 & 0.809 & 0.807 & 0.810 & --- \\
  Lifestyle only & 11 & 0.629 & 0.642 & 0.640 & 0.638 & 0.648 & $-$0.180 to $-$0.162 \\
  Baseline + lifestyle & 20 & 0.807 & 0.804 & 0.808 & 0.807 & 0.807 & $-$0.003 to 0.000 \\
  \bottomrule
  \end{tabular}}
\end{table*}

\begin{table*}[htbp]
\floatconts
  {tab:variability}
  {\caption{Variability of the lifestyle-minus-baseline held-out ROC-AUC difference over ten seeds. Training seeds keep the original development/test split and vary only the model and SMOTE random seeds. Repeated splits redraw the stratified 90/10 partition with ten seeds (including the original); because hyperparameters were tuned on the original development set, part of each new test set contributed to tuning, so these results describe split sensitivity conditional on that tuning, not independent validation. Each comparison has its own paired bootstrap 95\% CI; the last column counts intervals excluding zero.}}
  {\footnotesize\setlength{\tabcolsep}{4pt}\begin{tabular}{@{}lcccccc@{}}
  \toprule
  \bfseries Model & \bfseries Mean AUC (base / life) & \bfseries Mean $\Delta$ & \bfseries SD & \bfseries Range & \bfseries $\Delta > 0$ & \bfseries CI excl.\ 0 \\
  \midrule
  \multicolumn{7}{@{}l}{\itshape Training seeds, original split} \\
  \quad Logistic Regression & 0.809 / 0.807 & $-$0.0012 & 0.0002 & $-$0.0015 to $-$0.0009 & 0/10 & 0/10 \\
  \quad Random Forest & 0.803 / 0.804 & $+$0.0004 & 0.0010 & $-$0.0007 to $+$0.0018 & 5/10 & 0/10 \\
  \quad Gradient Boosting & 0.809 / 0.810 & $+$0.0012 & 0.0016 & $-$0.0008 to $+$0.0047 & 8/10 & 0/10 \\
  \quad Bagged Decision Tree & 0.806 / 0.808 & $+$0.0026 & 0.0014 & $+$0.0001 to $+$0.0042 & 10/10 & 0/10 \\
  \quad XGBoost & 0.809 / 0.809 & 0.0000 & 0.0017 & $-$0.0036 to $+$0.0026 & 6/10 & 0/10 \\
  \midrule
  \multicolumn{7}{@{}l}{\itshape Repeated random 90/10 splits (exploratory)} \\
  \quad Logistic Regression & 0.814 / 0.816 & $+$0.0022 & 0.0023 & $-$0.0018 to $+$0.0051 & 8/10 & 0/10 \\
  \quad Random Forest & 0.814 / 0.813 & $-$0.0009 & 0.0017 & $-$0.0028 to $+$0.0023 & 2/10 & 0/10 \\
  \quad Gradient Boosting & 0.816 / 0.817 & $+$0.0011 & 0.0022 & $-$0.0018 to $+$0.0043 & 6/10 & 0/10 \\
  \quad Bagged Decision Tree & 0.814 / 0.815 & $+$0.0004 & 0.0017 & $-$0.0024 to $+$0.0023 & 7/10 & 0/10 \\
  \quad XGBoost & 0.816 / 0.816 & $+$0.0002 & 0.0021 & $-$0.0028 to $+$0.0046 & 6/10 & 1/10 \\
  \bottomrule
  \end{tabular}}
\end{table*}
 
\section{Training-Seed and Repeated-Split Variability}
\label{apd:variability}
 
The primary analysis uses a single split and seed, so two checks
examine how much the lifestyle-minus-baseline difference varies
(Table~\ref{tab:variability}). First, keeping the original
development/test split fixed, both arms were refit with ten different
model and SMOTE random seeds. None of the 50 seed-by-model paired
intervals excluded zero, and mean differences ranged from $-$0.0012
(Logistic Regression) to $+$0.0026 (Bagged Decision Trees). Bagged
Decision Trees showed a positive difference in all ten seeds ($+$0.0001
to $+$0.0042), consistent in sign but small in size.
 
Second, as an exploratory check, the stratified 90/10 partition was
redrawn with ten seeds, including the original. Hyperparameters were
tuned on the original development set, so for the new seeds part of
each new test set contributed to tuning; this is split sensitivity
conditional on the original tuning, not independent validation. Mean
differences ranged from $-$0.0009 to $+$0.0022, and one of 50 paired
intervals excluded zero (XGBoost, seed 7: $+$0.0046 [$+$0.0006,
$+$0.0089]). Mean AUCs in both arms were somewhat higher on the new
splits than on the original test split (about 0.814--0.816 vs.\
0.804--0.810), which may reflect partition-to-partition variation, the
tuning overlap, or both.

\section{Reproducibility and Computational Details}
\label{apd:repro}

All code, result tables, figures, and raw SHAP arrays are available at
\url{https://github.com/urielbenymon/nhanes-lifestyle-hypertension}. The repository provides the full
execution order and maps every reported table and figure to the script
and output file that produce it. In brief, the main pipeline downloads
the NHANES files from CDC, builds the cohort and phenotypes, tunes and
freezes the models on the development set, and evaluates them on the
held-out test set; the robustness checks and all supplementary analyses
then reuse the frozen hyperparameters, and the appendix tables are
formatted directly from the saved result tables. The library-default
(untuned) results cited in Section~\ref{sec:methods} are saved
separately in the repository. A single random seed (42) was used
throughout for splitting, cross-validation folds, SMOTE, and all
stochastic model components, except in the training-seed,
repeated-split, and shuffled-lifestyle analyses, which vary seeds by
design. Table~\ref{tab:runtime} reports wall-clock
hyperparameter-search time per model (the dominant cost); all other
stages completed in seconds to low minutes. Analyses used Python 3.14
with pandas 3.0.3, numpy 2.4.6, scikit-learn 1.9.0, xgboost 3.4.1,
imbalanced-learn 0.14.2, shap 0.52.0, scipy 1.18.1, and matplotlib
3.11.1. Experiments were run on a Windows 11 system with an AMD Ryzen
Z1 Extreme processor (3.30~GHz) and 16~GB of installed RAM (12.7~GB
usable). No dedicated GPU was used.

\begin{table}[htbp]
\floatconts
  {tab:runtime}
  {\caption{Hyperparameter search wall-clock time per model (both arms combined).}}
  {\footnotesize\begin{tabular}{lr}
  \toprule
  \bfseries Model & \bfseries Total search time \\
  \midrule
  Logistic Regression & 3.7s \\
  Random Forest & 188.8s \\
  Gradient Boosting & 619.7s \\
  Bagged Decision Tree & 213.2s \\
  XGBoost & 116.5s \\
  \bottomrule
  \end{tabular}}
\end{table}

NHANES data were obtained from
\url{https://wwwn.cdc.gov/nchs/nhanes/continuousnhanes/default.aspx?Cycle=2017-2020}.

\section{Ethics, Scope, and Responsible-Use Notes}
\label{apd:ethics}

This is a secondary analysis of publicly available, de-identified NHANES
data; IRB status is stated in the Institutional Review Board statement
accompanying this submission. No clinical deployment is proposed or
validated. Reported associations describe predictive performance within
this analytic cohort; survey weights, primary sampling units, and strata
were not used, so estimates should not be interpreted as
population-representative or causal (Appendix~\ref{apd:cohort}).
Performance was not evaluated within sex, age, or race/ethnicity
subgroups, and fairness properties were not assessed; this is noted as
a limitation rather than an assumption of uniform performance across
demographic groups. The only subgroup analysis, restricted to
definitely untreated adults (Appendix~\ref{apd:untreated}), is an
exploratory sensitivity analysis that changes both the population and
the outcome definition; it is not a demographic subgroup or fairness
evaluation.
\end{document}